\documentclass[10pt,twocolumn,letterpaper]{article}

\usepackage[pagenumbers]{cvpr} %

\usepackage[dvipsnames]{xcolor}

\definecolor{cvprblue}{rgb}{0.21,0.49,0.74}
\usepackage[pagebackref,breaklinks,colorlinks,citecolor=cvprblue]{hyperref}

\usepackage{multirow}
\usepackage{tabularx}
\usepackage{graphicx}

\usepackage{bm}

\usepackage{amsmath}
	
\usepackage{wrapfig}
\usepackage{balance}

\usepackage{floatpag}

\newcommand{\boldstartspace}[1]{\vspace{0.1in}\noindent\textbf{#1}}

\title{MuRF: Multi-Baseline Radiance Fields}

\author{
Haofei Xu\textsuperscript{1,2} \quad 
Anpei Chen\textsuperscript{1,2} \quad 
Yuedong Chen\textsuperscript{3} \quad 
Christos Sakaridis\textsuperscript{1}  \quad 
Yulun Zhang\textsuperscript{4} \\
Marc Pollefeys\textsuperscript{1,5} \quad 
Andreas Geiger\textsuperscript{2,$\dagger$} \quad 
Fisher Yu\textsuperscript{1,$\dagger$} \\[3pt]
{\textsuperscript{1}ETH Zurich}
\,
{\textsuperscript{2}University of T\"ubingen, T\"ubingen AI Center}
\,
{\textsuperscript{3}Monash University}
\\
{\textsuperscript{4}Shanghai Jiao Tong University}
\,
{\textsuperscript{5}Microsoft}
\,
{\textsuperscript{$\dagger$}Joint last author}
\\[3pt]
\href{https://haofeixu.github.io/murf/}{haofeixu.github.io/murf}
}

\begin{document}

\maketitle

\begin{abstract}
We present Multi-Baseline Radiance Fields (MuRF), a general feed-forward approach to solving sparse view synthesis under multiple different baseline settings (small and large baselines, and different number of input views). To render a target novel view, we discretize the 3D space into planes parallel to the target image plane, and accordingly construct a target view frustum volume. Such a target volume representation is spatially aligned with the target view, which effectively aggregates relevant information from the input views for high-quality rendering. It also facilitates subsequent radiance field regression with a convolutional network thanks to its axis-aligned nature. The 3D context modeled by the convolutional network enables our method to synthesis sharper scene structures than prior works. Our MuRF achieves state-of-the-art performance across multiple different baseline settings and diverse scenarios ranging from simple objects (DTU) to complex indoor and outdoor scenes (RealEstate10K and LLFF). We also show promising zero-shot generalization abilities on the Mip-NeRF 360 dataset, demonstrating the general applicability of MuRF.

\end{abstract}

\setlength{\abovedisplayskip}{5pt}
\setlength{\belowdisplayskip}{5pt}

\vspace{-18pt}
\section{Introduction}
Novel view synthesis from sparse input views is a critical and practical problem in computer vision and graphics.
However, typical optimization-based Neural Radiance Fields (NeRFs)~\cite{mildenhall2020nerf,barron2021mip,barron2022mip,chen2022tensorf} cannot cope well with sparse views, and additional regularizations~\cite{niemeyer2022regnerf,wynn2023diffusionerf,truong2023sparf} are usually required to better constrain the optimization process.

In this paper, we aim at learning feed-forward NeRF models~\cite{yu2021pixelnerf,chen2021mvsnerf,wang2021ibrnet} that are able to perform feed-forward inference on new unseen data, \emph{without} requiring per-scene optimization. Moreover, inductive biases could be acquired from data and better results can potentially be attained.

\begin{figure}
    \centering

    \begin{subfigure}{0.49\linewidth}
        \includegraphics[width=\linewidth]{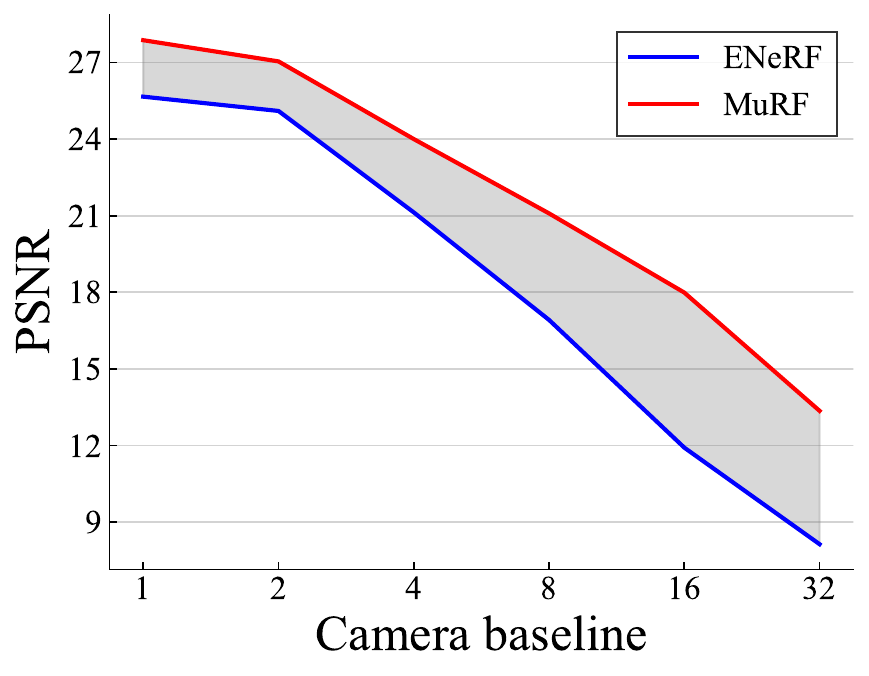}
        \caption{\scriptsize MuRF outperforms previous state-of-the-art small baseline method ENeRF. \quad The larger baseline, the larger performance gap.}
        \label{fig:teaser_small}
    \end{subfigure}
    \hfill
    \begin{subfigure}{0.49\linewidth}
        \includegraphics[width=\linewidth]{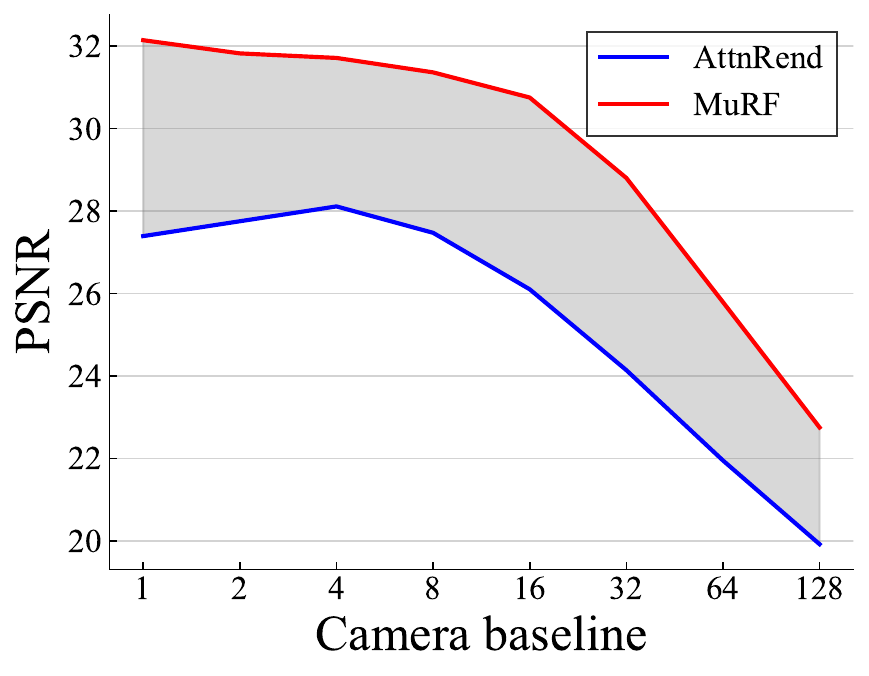}
        \caption{\scriptsize MuRF outperforms previous state-of-the-art large baseline method AttnRend. The smaller baseline, the larger performance gap.}
        \label{fig:teaser_large}
    \end{subfigure}

    \vspace{0.5em} %

    \begin{subfigure}{0.95\linewidth}
        \centering
        \includegraphics[width=\linewidth]{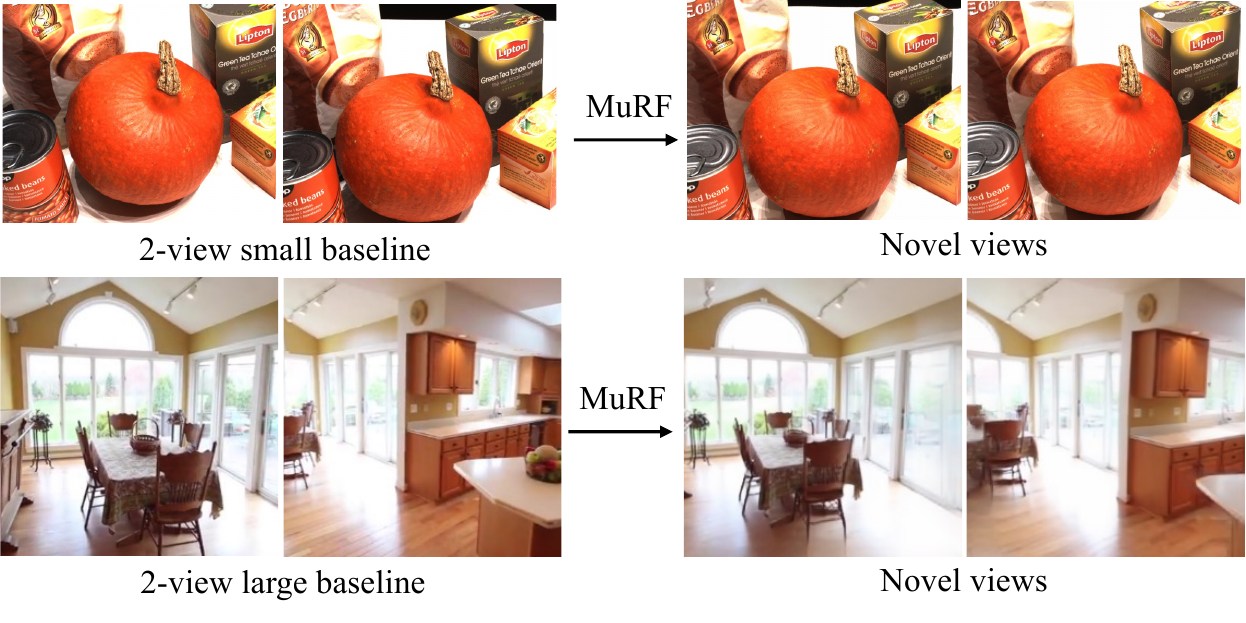}
        \label{fig:teaser_murf}
    \end{subfigure}
    \vspace{-8mm}

    \caption{\textbf{MuRF supports multiple different baseline settings}. Previous methods are specifically designed for either small (\eg, ENeRF~\cite{lin2022efficient}) or large (\eg, AttnRend~\cite{du2023learning}) baselines. However, no existing method performs well on both (see Table~\ref{tab:small_large}).} 
    
    \label{fig:teaser}
    \vspace{-5mm}
\end{figure}

To achieve this, existing sparse view methods can be broadly classified into methods that take small baseline images as input and methods that focus on large baseline settings. However, none of the existing methods works well across different baselines{\footnotemark} (see Fig.~\ref{fig:teaser_small} and Fig.~\ref{fig:teaser_large}).
In particular, small baseline methods~\cite{chen2021mvsnerf,wang2021ibrnet,chibane2021stereo,lin2022efficient,liu2022neural,johari2022geonerf,chen2023explicit} rely heavily on feature matching to extract relevant information from input views and suffer significantly when there is insufficient image overlap or occlusion occurs. 
In contrast, large baseline methods~\cite{sajjadi2022scene,du2023learning} are mostly data-driven, and often discard matching cues.

\footnotetext{We name the previously unnamed method of Du et al.~\cite{du2023learning} as AttnRend, according its \href{https://github.com/yilundu/cross_attention_renderer}{GitHub repository name}.}

Instead, they resort to large-scale datasets and generic architectures (\eg, Transformers~\cite{vaswani2017attention,dosovitskiy2020image}) to learn geometric inductive biases implicitly.
However, due to their correspondence-unaware and data-driven nature, even state-of-the-art methods~\cite{sajjadi2022scene,du2023learning} tend to produce blurry renderings and their generalization ability is still limited, as we show in Fig.~\ref{fig:sota_comparison} and Table~\ref{tab:generalization}.

Our goal is to address sparse view synthesis under both small and large camera baselines, different number of input views, and diverse scenarios ranging from simple objects to complex scenes.
To this end, we introduce Multi-Baseline Radiance Fields (MuRF), a method for sparse view synthesis from multiple different baseline settings (Fig.~\ref{fig:teaser}).

More specifically, to render a target novel view, we discretize the 3D space with planes parallel to the target image plane, and accordingly construct a \emph{target view frustum volume} that is spatially aligned with the target view to render. This is a crucial difference to previous volume-based methods like MVSNeRF~\cite{chen2021mvsnerf} and GeoNeRF~\cite{johari2022geonerf}, where the volumes are constructed in a \emph{pre-defined reference input view} by straightforwardly following the standard practice in multi-view stereo (MVS). However, a key difference between MVS and novel view synthesis (NVS) is that MVS estimates the depth of a \emph{reference input view}, while NVS aims to render a \emph{novel view}. Taking this difference into account is crucial, since the reference view volume will not be able to effectively aggregate information from input views when the overlap between the reference and target view is small, which results in failure for large baselines (Fig.~\ref{fig:visual_ablation_target_volume}). 

Our target view frustum volume is constructed by sampling multi-view input image colors and features, which provide appearance information to the prediction of the 3D point's color. We also compute the cosine similarities between sampled features to provide multi-view consistency cues to aid the prediction of volume density~\cite{chen2023explicit}. The sampled colors and features, and the computed cosine similarities are concatenated together to provide relevant information for both small and large baseline scenarios.

Equipped with this axis-aligned target volume representation, we further propose to use a convolutional neural network (CNN) to reconstruct the radiance field from this volume. Thanks to the context modeling abilities of CNNs, our method yields more accurate scene structures than previous (MLP-based) point-wise~\cite{chibane2021stereo,du2023learning} and (Ray Transformer-based) ray-wise methods~\cite{wang2021ibrnet,Varma2023ICLR,chen2023explicit} (see Fig.~\ref{fig:ablation_cnn_vs_mlp}).

In practice, we perform $8\times$ subsampling in the spatial dimension when constructing the target volume, which enables our method to efficiently handle high-resolution images. The full resolution radiance field is finally obtained with a lightweight $8 \times$ upsampler~\cite{shi2016real}. We also use a computationally more efficient 3D CNN alternative by factorizing the 3D $3 \times 3 \times 3$ convolution to a 2D $3 \times 3 \times 1$ convolution on the spatial dimension and a 1D $1 \times 1 \times 3$ convolution on the depth dimension, \ie, (2+1)D CNN, a popular strategy in video recognition works~\cite{tran2018closer,feichtenhofer2019slowfast}.

We conduct extensive experiments to demonstrate the effectiveness of our proposed target view frustum volume and 3D context-aware radiance field decoder. 
More specifically, we outperform previous specialized models on both small (DTU~\cite{jensen2014large}, LLFF~\cite{mildenhall2020nerf}) and large (RealEstate10K~\cite{DBLP:journals/tog/ZhouTFFS18}) baselines, and achieve promising zero-short generalization abilities on the Mip-NeRF 360~\cite{barron2022mip} dataset, indicating the general applicability of our method.

\vspace{-2mm}
\section{Related Work}
\vspace{-3mm}

\boldstartspace{Multi-Baseline}. The concept of ``multi-baseline" dates back to multi-baseline stereo depth estimation~\cite{OkutomiK93,gallup2008variable,yang2002real,yang2003multi}, where multiple stereo pairs with various baselines are used to surpass matching ambiguities and thus achieve improved precision. In this paper, we aim at designing a generally applicable approach to solving novel view synthesis from multiple different sparse view settings. When considering multiple images as input, our problem setup is somewhat similar to multi-baseline stereo, while with the radiance field as the output for volumetric rendering. Besides, we are also interested in novel view synthesis from very sparse (as few as 2) input views with different baselines. In this sense, our proposed multi-baseline radiance field not only draws connections to classic multi-baseline stereo, but also extends it to view synthesis from 2 input views with both small and large baselines. 

\begin{figure*}[t]
\centering
\includegraphics[width=\linewidth]{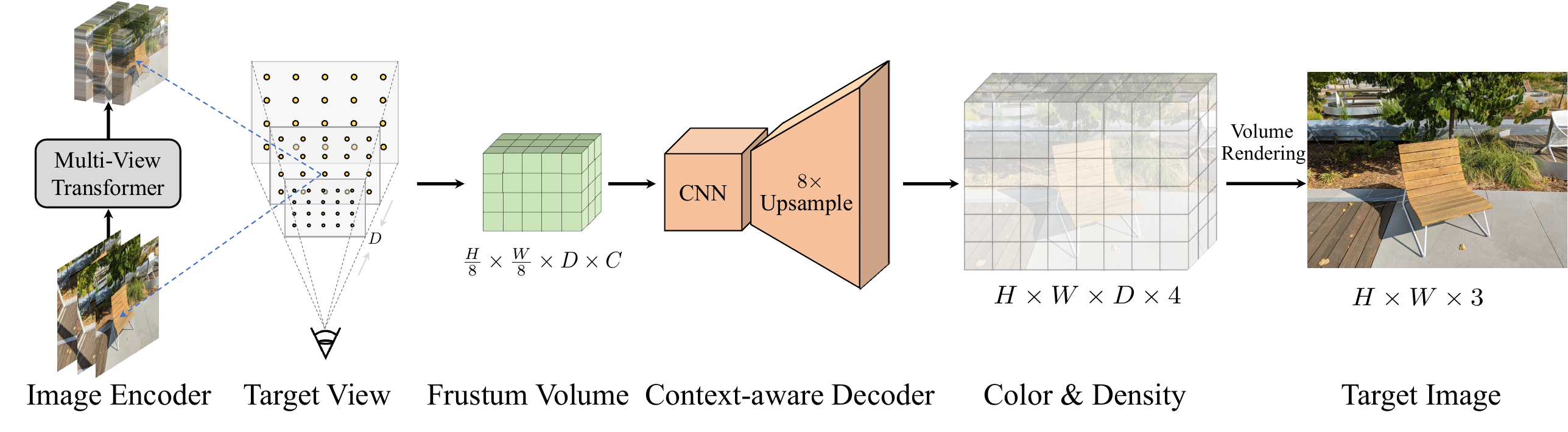}
\vspace{-8mm}
\caption{\textbf{Overview}. Given multiple input images, we first extract multi-view image features with a multi-view Transformer. To render a target image of resolution $H \times W$, we construct a target view frustum volume by performing $8\times$ subsampling in the spatial dimension while casting rays and sampling $D$ equidistant points on each ray. For each 3D point, we sample feature and color information from the extracted feature maps and input images, which consists of the elements of the target volume ${\bm z} \in \mathbb{R}^{\frac{H}{8} \times \frac{W}{8} \times D \times C}$. Here, $C$ denotes the channel dimension after aggregating sampled features and colors. To reconstruct the radiance field from the volume,  we model the context information in the decoder with a (2+1)D CNN operating on low resolution and subsequently obtain the full-resolution radiance field with a lightweight $8 \times$ upsampler. The target image is finally rendered with volumetric rendering.
}
 \label{fig:pipeline}
\vspace{-4.5mm}
\end{figure*}

\vspace{-2mm}
\boldstartspace{Sparse Input Novel View Synthesis}. Applying NeRF to novel view synthesis with sparse input views has garnered considerable interest in recent years. Several works improve the optimization-based pipeline via the incorporation of additional training regularizations, \eg, depth~\cite{niemeyer2022regnerf}, correspondence~\cite{truong2023sparf} and diffusion model~\cite{wynn2023diffusionerf}. An alternative area of research turns to the feed-forward designs by learning reliable representations from data, and accordingly removing the need of per-scene optimization. Among them, MVSNeRF~\cite{chen2021mvsnerf}, GeoNeRF~\cite{johari2022geonerf} and NeuRay~\cite{liu2022neural} follow the spirit of multi-view stereo (MVS) to construct cost volumes at pre-defined reference viewpoints. However, the reference volume will not be able to effectively aggregate the information for input views when the overlap between the reference and target novel view is small, which accordingly results in failure for large baselines. Besides, ENeRF~\cite{lin2022efficient} essentially relies on an MVS network to guide NeRF's sampling process with the estimated depth. However, the depth quality will be unreliable for large baselines and thus severely affects the subsequent rendering process.

Another line of works such as SRF~\cite{chibane2021stereo}, IBRNet~\cite{wang2021ibrnet}, GPNR~\cite{suhail2022generalizable}, GNT~\cite{Varma2023ICLR} and MatchNeRF~\cite{chen2023explicit} mostly perform ray-based rendering, where each ray is rendered independently and no explicit 3D context between different rays is modeled. In contrast, our target view volume representation effectively encodes the scene geometry and naturally enables 3D context modeling with a convolutional network, yielding better scene structures.
Despite various motivations and implementations, the above models are all designed and experimented only on small baseline settings, limiting their applications in real-world scenarios. Only a few existing approaches focus on the large baseline settings. Nevertheless, these works~\cite{sajjadi2022scene,du2023learning} resort to large-scale datasets and generic architectures, without employing explicit 3D representations, which ultimately leads to blurry rendering and limited generalization. As a comparison, our MuRF is developed with both geometry-aware target view frustum volume representation and 3D context-aware convolutional network, which make it excel at both small and large baseline settings on different datasets.

\vspace{-1mm}
\section{Approach}
\vspace{-1mm}

Our MuRF (Multi-Baseline Radiance Fields) is an encoder-decoder architecture (Fig.~\ref{fig:pipeline}), where the encoder maps the multi-view input images to features that are used to construct a volume at the target view's camera frustum, and the decoder regresses the radiance field from this volume representation. The target image is finally obtained using volume rendering~\cite{mildenhall2020nerf}. The key components are introduced below.

\subsection{Multi-View Feature Encoder}
\label{sec:encoder}
\vspace{-0.5mm}
To aggregate essential information required as input for learning feed-forward NeRF models, we first extract features $\{{\bm F}_k \}_{k=1}^K$ from $K$ input images $\{{\bm I}_k \}_{k=1}^K$. Our feature encoder consists of a weight-sharing per-image 2D CNN and a joint multi-view Transformer. The CNN consists of 6 residual blocks~\cite{he2016deep} and every 2 residual blocks include a stride-2 convolutional layer to achieve $2\times $ downsampling. Accordingly, we obtain features at $1/2$, $1/4$ and $1/8$ resolutions. The $1/8$ convolutional features are then fed into a joint multi-view Transformer to obtain features at $1/8$ resolution. Our Transformer is built on GMFlow~\cite{xu2022gmflow,xu2023unifying}'s 2-view Transformer architecture, where we extend the 2-view cross-attention to $K (K \geq 2)$ input views by performing cross-attention for all $K$ views simultaneously. This facilitates more efficient processing of additional input views without significantly increasing memory footprint compared to the pair-wise architecture~\cite{chen2023explicit}. The multi-scale features are sampled in next volume reconstruction step and the sampled features are concatenated together.

\subsection{Target View Frustum Volume}
\label{sec:target_volume}
\vspace{-0.5mm}
To render an image for a target novel view, our key difference with previous methods~\cite{chen2021mvsnerf,johari2022geonerf,liu2022neural} is that we construct our volume aligned with the \emph{target view frustum}, instead of a pre-defined reference input view. Such a spatially-aligned target view frustum volume representation naturally enables effective aggregation of information from input images. This is in contrast to the reference volume construction approach which has a high risk missing the information outside the epipolar lines of the reference view. This phenomenon is more pronounced for large baselines where the scene overlap is small, as illustrated in Fig.~\ref{fig:visual_ablation_target_volume}.

The volume elements are sampled from the input images and features to provide necessary cues to aid the prediction of color and density for volume rendering. More specifically, to render a target image of resolution $H \times W$, we perform $8 \times $ subsampling in the spatial dimension while casting rays and uniformly sample $D$ points on each ray. The $8\times$ subsampling enables our method to maintain an acceptable volume resolution for high-resolution images, and we finally obtain the full resolution volume with a lightweight upsampler~\cite{shi2016real}. Next, we introduce the sampling process.

\boldstartspace{Color sampling within a window.} Due to the $8 \times$ subsampling, we might risk losing some information if we only sample a single point compared to full resolution sampling. To remedy this, we instead sample a $9\times 9$ window centered at the 2D projection of the 3D point and thus obtain a color vector of $\{\tilde{\bm c}_k^{\mathrm{w}} \}_{k=1}^K$ for $K$ input views, where $\tilde{\bm c}_k^{\mathrm{w}}$ denotes the concatenation of colors within the window. 

\vspace{-1.5mm}
\boldstartspace{Feature sampling and matching.} We also sample image features $\{{\bm f}_i \}_{i=1}^K$ from $K$ feature maps. To obtain geometric cues, we quantify the multi-view consistency of these features by computing their pair-wise cosine similarities. We then use the cosine feature similarities to provide geometric cues for volume density prediction, based on the observation that the multi-view features of a surface point tend to exhibit high multi-view consistency~\cite{chen2023explicit}. All the pair-wise cosine similarities are aggregated with a learned weighted average, which can be expressed as
\begin{equation}
\label{eq:cosine}
    \hat{\bm s} = w_{ij} \sum_{i < j} \cos({\bm f}_i, {\bm f}_j), \quad i, j = 1, 2, \ldots, K
\end{equation}
where ${\bm s}_{ij}$ denotes the cosine similarity between pair $({\bm f}_i, {\bm f}_j)$, and $(i, j)$ iterates over all possible $K(K-1)/2$ non-repeated pairs. Following previous work~\cite{zhang2020visibility}, $w_{ij}$ are normalized visibility weights learned from the entropy of the depth distribution on each ray.

It's worth noting that the pair-wise cosine similarities $\cos({\bm f}_i, {\bm f}_j)$ in Eq.~\eqref{eq:cosine} can be computed in parallel with a single matrix multiplication. More specifically, we first normalize the features to unit vectors and then collect them as a $K \times M$ matrix, where $M$ is the feature dimension. Finally all $K(K-1)/2$ pair-wise cosine similarities can be obtained by extracting the upper triangle of the multiplication of this matrix with its transpose. In practice, we use group-wise cosine similarities~\cite{guo2019group,chen2023explicit} for further improved expressiveness, where the same computation still applies.

\vspace{-1.5mm}
\boldstartspace{Multi-view aggregation.} Next, the sampled colors and features from $K$ input views are aggregated with a learned weighted average. More specifically, the aggregated color and features are
\begin{equation}
\label{eq:color_feature}
    \hat{\bm c} = w_{i} \sum_{i=1}^K \tilde{\bm c}_i^{\mathrm{w}}, \quad \hat{\bm f} = w_{i} \sum_{i=1}^K {\bm f}_i,
\end{equation}
where $w_{i}$ are normalized learned weights~\cite{lin2022efficient}. In particular, we use a small MLP network to predict $w_{i}$ from the concatenation of sampled features and view direction difference between the source and target views.

The cosine similarity $\hat{\bm s}$ in Eq.~\eqref{eq:cosine}, colors $\hat{\bm c}$ and features $\hat{\bm f}$ in Eq.~\eqref{eq:color_feature} are then concatenated and projected to a $C$-dimensional vector with a linear layer. Accordingly we obtain a target volume ${\bm z} \in \mathbb{R}^{\frac{H}{8} \times \frac{W}{8} \times D \times C}$ for all $\frac{H}{8} \times \frac{W}{8} \times D$ 3D points. This volume encodes appearance and geometry information from multi-view images and features, which is next fed as input to the decoder for radiance field prediction.

\subsection{Context-aware Radiance Field Decoder}
\label{sec:decoder}

Given the target view frustum volume ${\bm z} \in \mathbb{R}^{\frac{H}{8} \times \frac{W}{8} \times D \times C}$, our decoder learns to predict the 4-dimensional (color and density) radiance field ${\bm R} \in {\mathbb{R}^{H \times W \times D \times 4}}$ of all $H \times W \times D$ 3D points. In order to perform well on both small and large baseline input views, we observe that it's crucial to model the 3D context between different 3D points. The context information can help learn useful inductive biases from data and accordingly leads to better scene structures.

To achieve this, we model the 3D context with a convolutional network. A straightforward approach would be using a 3D CNN. In this paper, we explore an alternative for better memory and parameter efficiency, while maintaining similar performance (Fig.~\ref{tab:ablation}). More specifically, we factorize the 3D ($3 \times 3 \times 3$) convolution to a 2D ($3 \times 3 \times 1$) convolution in the spatial dimension and a 1D ($1 \times 1 \times 3$) convolution in the depth dimension, \ie, (2+1)D CNN, which is a popular strategy in video recognition works~\cite{tran2018closer,feichtenhofer2019slowfast}.

The full decoder consists of two major components: a (2+1)D CNN which operates on the low resolution volume and a lightweight upsampler which achieves $8 \times $ upsampling and thus outputs a full-resolution radiance filed. 
The low-resolution (2+1)D CNN architecture is composed of 12 stacked (2+1)D residual blocks~\cite{he2016deep}. 
The final color and density predictions are obtained using two linear layers, with output channels of 3 and 1, respectively. 
More architectural details are presented in the supplementary material.

\subsection{Hierarchical Volume Sampling}

Like other NeRF methods~\cite{mildenhall2020nerf,wang2021ibrnet,barron2021mip}, our model also supports hierarchical volume sampling for further improved performance. When hierarchical sampling is used, the model presented before can be viewed as a coarse model, and the hierarchical stage as a fine model. The fine model has a very similar overall architecture as the coarse stage, with the ray sampling process as the key difference. More specifically, given the density prediction from the coarse model, we first compute the Probability Distribution Function (PDF)~\cite{mildenhall2020nerf} on each ray by normalizing the color composition weights in the volume rendering equation. Next, we sample a new set of points on each ray according to this distribution. Thanks to the coarse geometry predicted by the coarse network, the fine model requires sampling fewer points since the influence of empty space or occluded regions can be removed. In our implementation, we uniformly sample 64 points on each ray in the coarse model, and only sample 16 points in the fine stage. Due to the smaller number of sampling points, we are able to directly construct a target view frustum volume at the full resolution. We also reduce the volume's channel dimension from 128 to 16, which makes it more efficient for subsequent regression. Thus the volume at the fine stage is more compact. We use a lightweight 3D U-Net~\cite{ronneberger2015,rombach2022high} to predict the color and density from this volume, and the final color prediction is similarly obtained using volume rendering~\cite{mildenhall2020nerf}. 
More details are presented in the supplementary material.

\subsection{Training Loss}

We use random crops from the full image for training, the training loss is an addition of $\ell_1$, SSIM and LPIPS losses between rendered and ground truth image color.

\vspace{-1.5mm}
\section{Experiments}
\vspace{-3mm}

\boldstartspace{Implementation details}. We implement our method using PyTorch~\cite{paszke2019pytorch} and we adopt a two-stage training process, where we first train the coarse model only, and then train the fine model with the coarse model frozen. For all the experiments, we sample 64 points in the coarse model, and 16 points in the fine model. For experiments on the RealEstate10K~\cite{DBLP:journals/tog/ZhouTFFS18} dataset, we follow AttnRend~\cite{du2023learning}'s setting to train and evaluate on the $256\times 256$ resolution. On this resolution, our method can easily use $4\times$ subsampling when constructing the volume, unlike the default $8 \times$ subsampling. We also reduce the number of channels of the volume from 128 to 64 when $4\times $ subsampling is used. We don't include the additional hierarchical sampling stage for experiments on the RealEstate10K dataset since the single scale model already performs very competitively with $4\times$ subsampling. Our code 
is available at \url{https://github.com/autonomousvision/murf}.

\vspace{-1.5mm}
\boldstartspace{Evaluation settings}.
Our evaluations include both small and large baselines, different number of views, and diverse scenarios, including object-centric, indoor and unbounded outdoor scenes, to test the model's general applicability.

\subsection{Main Results}
\vspace{-3mm}
\boldstartspace{Both small \& large baselines}. To compare with existing methods on \emph{both} small and large baselines, we choose previous state-of-the-art small baseline method ENeRF~\cite{lin2022efficient} and large baseline method AttnRend~\cite{du2023learning} as two representative approaches. The evaluations are conducted on DTU~\cite{jensen2014large} and RealEstate10K~\cite{DBLP:journals/tog/ZhouTFFS18} datasets. Since ENeRF only reported its results on DTU and AttnRend only on RealEstate10K, not both, we re-train ENeRF on RealEstate10K and AttnRend on DTU, which allows for more comprehensive comparisons. The results in Table~\ref{tab:small_large} demonstrate that previous state-of-the-art methods are specialized to either small or large baselines, but they can not work well on both. In contrast, our MuRF performs consistently better on both small and large baselines. Next, we conduct comparisons in different specialized settings.

\begin{figure}[t]
\centering
\includegraphics[width=\linewidth]{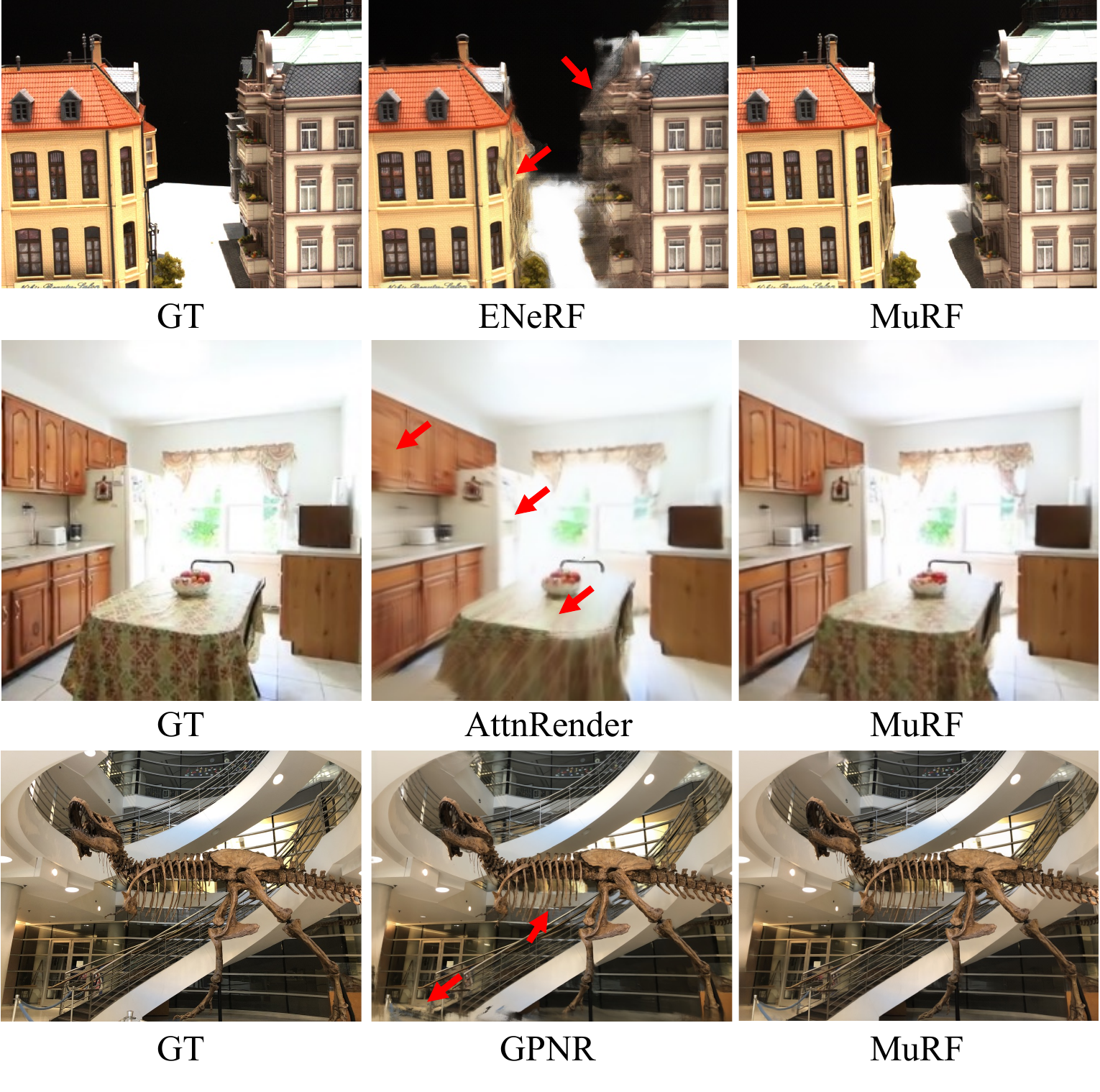}
\vspace{-9mm}
\caption{\textbf{Visual comparisons with previous best methods} on DTU, RealEstate10K and LLFF datasets. 
}
 \label{fig:sota_comparison}
\vspace{-6.5mm}
\end{figure}

\vspace{-1.5mm}
\boldstartspace{3-view small baseline on DTU}. In this setting, to render a target view, 3 nearest views are selected from all source views based on the distance to the target camera location. Thus this constitutes a small baseline setting. As shown in Table~\ref{tab:dtu_3view_small}, we achieve more than 1dB PSNR improvement compared to previous best method ENeRF~\cite{lin2022efficient}. The visual comparisons are shown in Fig.~\ref{fig:sota_comparison}, where our method produces significantly better scene structures.

\begin{table}[t]
\vspace{1mm}
    \begin{center}
\footnotesize
    \setlength{\tabcolsep}{2.5pt} %
    \resizebox{\linewidth}{!}{
    \begin{tabular}{lcccccccccccccccccccccccc}
    \toprule
    \multirow{2}{*}[-2pt]{Method} & \multicolumn{3}{c}{DTU (small baseline)} & \multicolumn{3}{c}{RealEstate10K (large baseline)} \\
    \addlinespace[-12pt] \\
    \cmidrule(lr){2-4} \cmidrule(lr){5-7} 
    \addlinespace[-12pt] \\
    & PSNR$\uparrow$ & SSIM$\uparrow$ & LPIPS$\downarrow$ & PSNR$\uparrow$ & SSIM$\uparrow$ & LPIPS$\downarrow$ \\
    
    \midrule

    ENeRF~\cite{lin2022efficient} & 27.61 & 0.956 & 0.091 & 19.52 & 0.739 & 0.341 \\

    AttnRend~\cite{du2023learning} & 18.57 & 0.732 & 0.419 & 21.38 & 0.839 & 0.262 \\
    
    MuRF  & \textbf{28.76} & \textbf{0.961} & \textbf{0.077} & \textbf{24.20} & \textbf{0.865} & \textbf{0.170} \\
    
    \bottomrule
    \end{tabular}
    }
    \end{center}
    \vspace{-6mm}
    \caption{\textbf{Comparison on both small and large baselines}. Previous state-of-the-art methods are specialized to either small (ENeRF) or large (AttnRend) baselines, but can not work well on both. 
    }
    \label{tab:small_large}
    \vspace{-3mm}

\end{table}

\begin{table}[!t]
\begin{center}
\vspace{1.5mm}
\small
\begin{tabular}{lcccccccccccccccccccccccc}
\toprule
Method & PSNR$\uparrow$ & SSIM$\uparrow$ & LPIPS$\downarrow$  \\
    
\midrule

    PixelNeRF~\cite{yu2021pixelnerf}                                     & 19.31          & 0.789          & 0.382             \\
    SRF~\cite{chibane2021stereo} & 22.12 & 0.845 & 0.292 \\
    IBRNet~\cite{wang2021ibrnet}                                        & 26.04          & 0.917          & 0.190         \\
    MVSNeRF~\cite{chen2021mvsnerf}                                       & 26.63  & 0.931  & 0.168     \\
    GeoNeRF~\cite{johari2022geonerf} &  26.76 & 0.893 & 0.150 \\
    MatchNeRF~\cite{chen2023explicit}                              & 26.91 & 0.934 & 0.159    \\
    ENeRF~\cite{lin2022efficient} & 27.61 & 0.956 & 0.091 \\
    MuRF  & \textbf{28.76} & \textbf{0.961} & \textbf{0.077} \\

    \bottomrule
    \end{tabular}
    \end{center}
    \vspace{-6mm}
    \caption{\textbf{DTU 3-view small baseline}. 
    }
    \label{tab:dtu_3view_small}
    \vspace{-3mm}

\end{table}

\begin{table}[!t]
\begin{center}
\small
    \begin{tabular}{lcccccccccccccccccccccccc}
    \toprule
    Method & PSNR$\uparrow$ & SSIM$\uparrow$ & LPIPS$\downarrow$  \\
    
    \midrule

    PixelNeRF~\cite{yu2021pixelnerf} & 13.91 & 0.460 & 0.591 \\
    SRF~\cite{chibane2021stereo} & 15.40 & 0.486 & 0.604 \\
    GeoNeRF~\cite{johari2022geonerf} & 16.65 & 0.511 & 0.541 \\
    IBRNet~\cite{wang2021ibrnet} & 15.99 & 0.484 & 0.532 \\
    GPNR~\cite{suhail2022generalizable} & 18.55 & 0.748 & 0.459 \\
    AttnRend~\cite{du2023learning} & 21.38 & 0.839 & 0.262 \\
    MuRF  & \textbf{24.20} & \textbf{0.865} & \textbf{0.170} \\

    \bottomrule
    \end{tabular}
    \end{center}
    \vspace{-6mm}
    \caption{\textbf{RealEstate10K 2-view large baseline}. 
    }
    \label{tab:realestate}
    \vspace{-5mm}

\end{table}

\begin{table*}[t]
    \begin{center}
\footnotesize
    \setlength{\tabcolsep}{2.5pt} %
    \begin{tabular}{lcccccccccccccccc}
    \toprule
    
    \multirow{2}{*}[-2pt]{Method} & \multirow{2}{*}[-2pt]{Setting} & \multicolumn{3}{c}{3-view} & \multicolumn{3}{c}{6-view} & \multicolumn{3}{c}{9-view} \\
    \addlinespace[-12pt] \\
    \cmidrule(lr){3-5} \cmidrule(lr){6-8} \cmidrule(lr){9-11} 
    \addlinespace[-12pt] \\
    & & PSNR$\uparrow$ & SSIM$\uparrow$ & LPIPS$\downarrow$ & PSNR$\uparrow$ & SSIM$\uparrow$ & LPIPS$\downarrow$ & PSNR$\uparrow$ & SSIM$\uparrow$ & LPIPS$\downarrow$ \\
    \midrule
    Mip-NeRF~\cite{barron2021mip} & \multirow{5}{*}[-2pt]{\begin{tabular}[x]{@{}c@{}}per-scene \\ optimization \end{tabular}} & 8.68 & 0.571 & 0.353 & 16.65 & 0.741 & 0.198 & 23.58 & 0.879 & 0.092  \\
    DietNeRF~\cite{jain2021putting} & & 11.85 & 0.633 & 0.314 & 20.63 & 0.778 & 0.201 & 23.83 & 0.823 & 0.173 \\
    RegNeRF~\cite{niemeyer2022regnerf} & & 18.89 & 0.745 & 0.190 & 22.20 & 0.841 & 0.117 & 24.93 & 0.884 & 0.089 \\
    DiffusioNeRF~\cite{wynn2023diffusionerf} & & 16.20 & 0.698 & 0.207 & 20.34 & 0.818 & 0.139 & 25.18 & 0.883 & 0.095  \\
    SPARF~\cite{truong2023sparf} & & 21.01 & 0.870 & \textbf{0.100} & - & - & - & - & - & - & \\
    
    \midrule

    SRF~\cite{chibane2021stereo} &  \multirow{5}{*}[-2pt]{\begin{tabular}[x]{@{}c@{}} feed-forward \\ inference \end{tabular}} & 15.32 & 0.671 & 0.304 & 17.54 & 0.730 & 0.250 & 18.35 & 0.752 & 0.232 \\
    PixelNeRF~\cite{yu2021pixelnerf} & & 16.82 & 0.695 & 0.270 & 19.11 & 0.745 & 0.232 & 20.40 & 0.768 & 0.220 \\
    MVSNeRF~\cite{chen2021mvsnerf} & & 18.63 & 0.769 & 0.197 & 20.70 & 0.823 & 0.156 & 22.40 & 0.853 & 0.135 \\
    ENeRF$^\dagger$~\cite{lin2022efficient} & & 19.84 & 0.856 & 0.171 & 21.82 & 0.894 & 0.131 & 23.49 & 0.919 & 0.106 \\
    MuRF & & \textbf{21.31} & \textbf{0.885} & 0.127 & \textbf{23.74} & \textbf{0.921} & \textbf{0.095} & \textbf{25.28} & \textbf{0.936} & \textbf{0.084} \\

    \bottomrule
    \end{tabular}
    \end{center}
    \vspace{-6mm}
    \caption{\textbf{DTU 3, 6 and 9 input views}. $^\dagger$Enhanced ENeRF baseline by doubling the number of sampling points on each ray, otherwise we were not able to obtain meaningful results. We again outperform ENeRF by 1$\sim$2dB PSNR even compared to this enhanced baseline.
    }
    \label{tab:regnerf_dtu}
    \vspace{-4.5mm}
\end{table*}

\vspace{-1.5mm}
\boldstartspace{2-view large baseline on RealEstate10K}. The RealEstate10K is a very large dataset, which consists of more than 66K training scenes and more than 7K testing scenes. We follow the evaluation setting of AttnRend~\cite{du2023learning} for comparisons. More specifically, the two input views are selected from a video with a distance of 128 frames, and the target view to synthesis is an intermediate frame. This is a challenging setting since the overlap between two input views is usually small, useful priors should be acquired from the model and the data in order to obtain reliable synthesis results. In this large baseline setting, we achieve significant improvement ($\sim$3dB PSNR) than previous best method AttnRend~\cite{du2023learning}, as shown in Table~\ref{tab:realestate}. The visual comparison in Fig.~\ref{fig:sota_comparison} indicates that our method produces clearer rendering results than AttnRend~\cite{du2023learning}. We attribute such large improvements to our target view frustum volume and the convolutional radiance field decoder, which are two key missing components in AttnRend~\cite{du2023learning}. The target view frustum volume effectively encodes the geometric structure of the view synthesis task and the convolutional decoder enables context-modeling between neighbouring 3D points, both significantly contributing to our final performance. Their effects are more thoroughly analyzed in Sec.~\ref{sec:ablation}.

\vspace{-1.5mm}
\boldstartspace{Different number of input views on DTU and LLFF}. We also evaluate on different number of input views to further understand the effect of different baselines. Firstly, we follow RegNeRF~\cite{niemeyer2022regnerf}'s setting of 3, 6 and 9 input views on DTU. Different from the aforementioned 3-view small baseline setting on DTU, the baseline of this setting is larger. More specifically, for each test scene, different numbers of views (3, 6 and 9) are sampled from a fixed set of 9 views. Each scene has 25 test views and they all share the same input views. This is different from the small baseline DTU setting before where the nearest views (instead of shared fixed views for all test views in each scene) are selected for each test view. Thus this setting is more close to the large baseline setting, especially when the number of input views is small. In this setting, our MuRF significantly outperforms previous representative methods like PixelNeRF~\cite{yu2021pixelnerf} and MVSNeRF~\cite{chen2021mvsnerf}, as shown in Table~\ref{tab:regnerf_dtu}.

\begin{table}[!t]
    \begin{center}
\footnotesize
    \begin{tabular}{lcccccccccccccccccccccccc}
    \toprule
    Method & \#views & PSNR$\uparrow$ & SSIM$\uparrow$ & LPIPS$\downarrow$  \\
    
    \midrule

    PixelNeRF~\cite{yu2021pixelnerf} & 10 & 18.66 & 0.588 & 0.463 \\
    MVSNeRF~\cite{chen2021mvsnerf} & 10 & 21.18 & 0.691 & 0.301 \\
    IBRNet~\cite{wang2021ibrnet} & 10 & 25.17 & 0.813 & 0.200 \\
    NeuRay~\cite{liu2022neural} & 10 & 25.35 & 0.818 & 0.198 \\
    GPNR~\cite{suhail2022generalizable} & 10 & 25.72 & 0.880 & 0.175 \\
    GNT~\cite{wang2022attention} & 10 & 25.53 & 0.885 & 0.218 \\
    \midrule
    \multirow{3}{*}[-2pt]{MuRF}  & 4 & 25.95 & 0.897 & 0.149 \\
    & 6 & 26.04 & 0.900 & 0.153 \\
     & 10 & \textbf{26.49}& \textbf{0.909} & \textbf{0.143} \\
     
     \bottomrule
    \end{tabular}
    \end{center}
    \vspace{-6mm}
    \caption{\textbf{LLFF}. Our 4-view model already outperforms previous 10-view methods. It improves further with more views.
    }
    \label{tab:llff}
    \vspace{-7mm}

\end{table}

\begin{figure*}[!t]
    \centering
    \includegraphics[width=\linewidth]{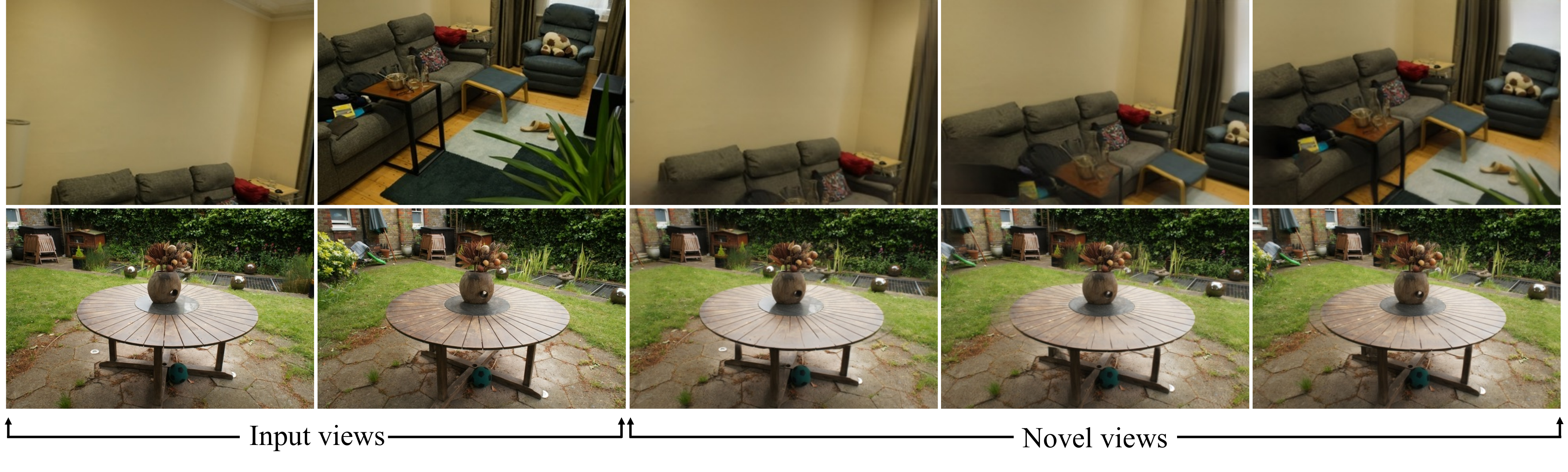}
    \vspace{-9mm}
    \caption{\textbf{Zero-shot generalization on Mip-NeRF 360 dataset}. 
    }
    \label{fig:zero_mipnerf360}
    \vspace{-2mm}
\end{figure*}

\begin{figure*}[!t]
    \centering
    \includegraphics[width=\linewidth]{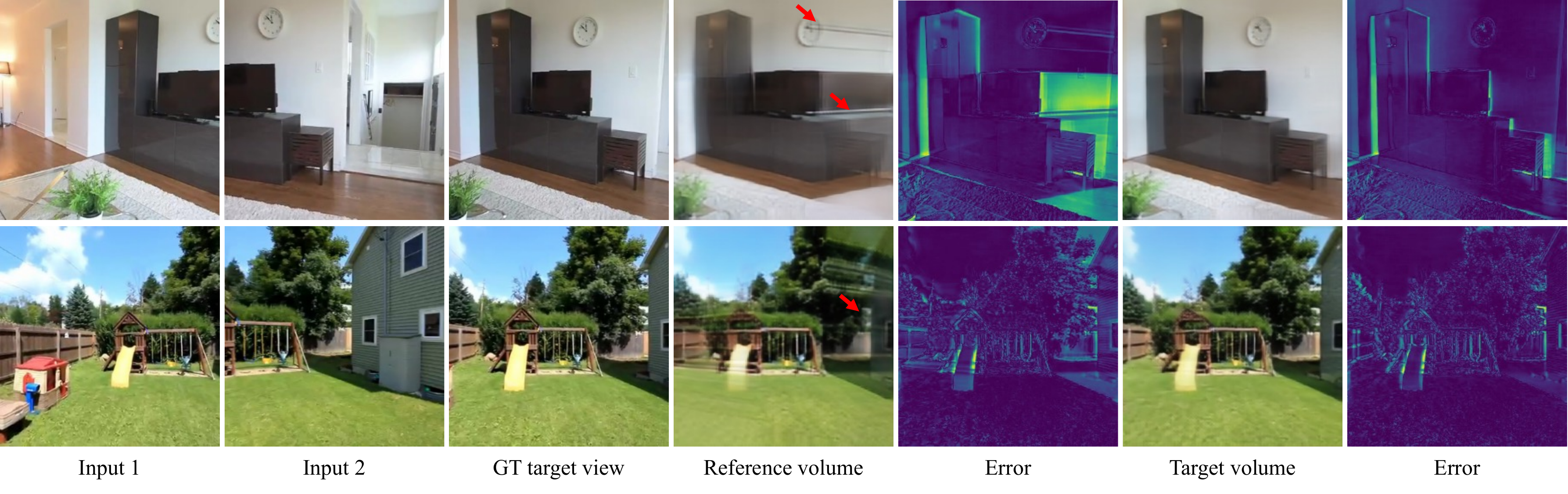}
    \vspace{-8mm}
    \caption{\textbf{Our target view volume \vs. reference (first image) view volume}. Constructing the volume at the pre-defined reference view space might miss relevant information (\eg, red arrows regions) in other views since such information could be far away from the the reference view's epipolar lines and thus hard to be sampled. In contrast, we construct the volume in the target view, which more effectively aggregates information from all input views and thus maximizes the information usage.
    }
    \label{fig:visual_ablation_target_volume}
    \vspace{-4.5mm}
\end{figure*}

To better understand the performance of different methods on different baselines, we re-train ENeRF~\cite{lin2022efficient}, the previous state-of-the-art method on DTU's small baseline setting,  on this large baseline setting with their official code. Since the key component of ENeRF is to use depth-guided sampling to skip empty space in NeRF's volume rendering process, its performance heavily relies on the quality of the estimated depth. In this large baseline setting, our re-trained ENeRF is not able to produce reasonable results with their original hyper-parameters since the small overlap between input views makes the depth estimation quality lower and thus negatively affect NeRF's rendering process. Thus we train an enhanced ENeRF baseline by doubling the number of sampling points (from original 8 and 2 to 16 and 4 for 2-stage NeRFs) on each ray such that the model is more tolerant to the depth estimation error. The results are shown in Table~\ref{tab:regnerf_dtu}. Our MuRF again outperforms ENeRF by 1$\sim$2dB PSNR in different number of input views.

Compared with per-scene optimization methods, our MuRF achieves similar performance with the state-of-the-art 3-view method SPARF~\cite{truong2023sparf} in the 3-view setting. It's worth noting that SPARF specifically focuses on 3 input views and they didn't report the performance on 6 and 9 views. In contrast, we purse thorough evaluation on different number of views and we show clear improvement than previous per-scene optimization methods with 6 and 9 input views, despite without using any per-scene optimization.

We also compare with previous methods on the LLFF dataset~\cite{mildenhall2020nerf}. Since this is a forward-facing dataset, the baselines between different images are typically small. Previous methods usually evaluate on this dataset with 10 input views following IBRNet~\cite{wang2021ibrnet}'s setting. Surprisingly, our MuRF trained with 4 input views already outperforms all previous 10-view methods (Table~\ref{tab:llff}), suggesting our model is also more data-efficient. Our performance gets further improvement with more input views. The visual comparison with previous state-of-the-art method GPNR~\cite{suhail2022generalizable} in Fig.~\ref{fig:sota_comparison} shows that our rendering has better structures.

\begin{table}[t]
\vspace{1mm}
    \begin{center}
\footnotesize
    \setlength{\tabcolsep}{2.5pt} %
    \resizebox{\linewidth}{!}{
    \begin{tabular}{lcccccccccccccccccccccccc}
    \toprule
    \multirow{2}{*}[-2pt]{Method} & \multicolumn{3}{c}{DTU} & \multicolumn{3}{c}{Mip-NeRF 360 Dataset} \\
    \addlinespace[-12pt] \\
    \cmidrule(lr){2-4} \cmidrule(lr){5-7} 
    \addlinespace[-12pt] \\
    & PSNR$\uparrow$ & SSIM$\uparrow$ & LPIPS$\downarrow$ & PSNR$\uparrow$ & SSIM$\uparrow$ & LPIPS$\downarrow$ \\
    
    \midrule

    AttnRend~\cite{du2023learning} & 11.35 & 0.567 & 0.651 & 14.00 & 0.474 & 0.712 \\
    MuRF & \textbf{22.19} & \textbf{0.894} & \textbf{0.211} & \textbf{23.98} & \textbf{0.800} & \textbf{0.293} \\
    
    \bottomrule
    \end{tabular}
    }
    \end{center}
    \vspace{-6.5mm}
    \caption{\textbf{Zero-shot generalization after trained on RealEstate10K}. Our method outperforms AttnRend~\cite{du2023learning} by significantly large margins.
    }
    \label{tab:generalization}
    \vspace{-6.5mm}

\end{table}

\begin{table*}[!t]
    \begin{center}
\footnotesize
    \begin{tabular}{llccccccccccccccccccccccc}
    \toprule
    \multirow{2}{*}[-2pt]{Module} & \multirow{2}{*}[-2pt]{Method} &  \multicolumn{3}{c}{Large Baseline (RealEstate10K)} & \multicolumn{3}{c}{Small Baseline (DTU)} \\ %
    \addlinespace[-12pt] \\
    \cmidrule(lr){3-5} \cmidrule(lr){6-8} 
    \addlinespace[-12pt] \\
    & & PSNR$\uparrow$ & SSIM$\uparrow$ & LPIPS$\downarrow$ & PSNR$\uparrow$ & SSIM$\uparrow$ & LPIPS$\downarrow$  \\
    
    \midrule

    \multirow{2}{*}[-2pt]{Volume Orientation} & \underline{target view} & \textbf{21.29} & \textbf{0.808} & \textbf{0.246} & \textbf{26.89} & \textbf{0.925} & \textbf{0.119} \\ %
    & reference view & 20.20 & 0.776 & 0.309 & 26.50 & 0.922 & 0.138  \\

    \midrule

    \multirow{4}{*}[-2pt]{Volume Elements} & \underline{cosine \& feature \& color } & \textbf{21.29} & \textbf{0.808} & \textbf{0.246} & \textbf{26.89} & \textbf{0.925} & \textbf{0.119} \\ %
    & w/o cosine & 21.06 & 0.800 & 0.255  & 26.68 & 0.919 & 0.124 \\ %
    & w/o feature & 20.91 & 0.794 & 0.275 & 26.79 & 0.923 & 0.132 \\ %
    & w/o color & 20.04 & 0.789 & 0.275 & 25.80 & 0.916 & 0.133 \\ %

    \midrule

    \multirow{6}{*}[-2pt]{Radiance Decoder} & \underline{(2+1)D CNN} (spatial + depth) & 21.29 & 0.808 & 0.246 & \textbf{26.89} & \textbf{0.925} & \textbf{0.119} \\ %
    & 3D CNN (spatial \& depth) & \textbf{21.39} & \textbf{0.809} & \textbf{0.243} & 26.86 & 0.924 & 0.121 \\ %
    & 2D CNN (spatial only) & 21.06 & 0.801 & 0.256 & 26.67 & 0.921 & 0.124 \\ %
    & 1D CNN (depth only) & 20.78 & 0.788 & 0.272 & 26.37 & 0.920 & 0.124 \\ %
    & Ray Transformer (depth only) & 20.57 & 0.780 & 0.283 & 26.22 & 0.912 & 0.129 \\ %
    & MLP (point only) & 20.41 & 0.778 & 0.298 & 26.24 & 0.918 & 0.126 \\

    \bottomrule
    \end{tabular}
    \end{center}
    \vspace{-6mm}
    \caption{\textbf{Ablations}. Settings used in our final model are underlined. The 3D CNN model has more than 50\% parameters (16.6M \vs  10.4M) than our factorized (2+1)D CNN model, and the performance is similar, thus we choose to use the (2+1)D CNN.
    }
    \label{tab:ablation}
    \vspace{-5mm}

\end{table*}

\vspace{-1.5mm}
\boldstartspace{Generalization on DTU and Mip-NeRF 360 dataset}. For feed-forward models, we are also interested in their zero-shot generalization abilities on unseen datasets, which is a practical problem setting. For this evaluation, we compare with AttnRend~\cite{du2023learning} using their released model trained on the same RealEstate10K dataset. The results are reported on the object-centric dataset DTU and scene-level Mip-NeRF 360 dataset with 2 input views. From Table~\ref{tab:generalization}, we observe our MuRF generalizes significantly better than AttnRend.
The lack of geometric inductive biases makes AttnRend more data-hungry and prone to overfitting to the training data.

To further explore the zero-shot performance limit on Mip-NeRF 360 dataset, we conduct additional fine-tuning with our RealEstate10K pre-trained model on a mixed dataset collected by IBRNet~\cite{wang2021ibrnet}. We achieve further improvement and the promising visual results shown in Fig.~\ref{fig:zero_mipnerf360} indicate the general applicability of our method.

\subsection{Ablation and Analysis}
\label{sec:ablation}
\vspace{-2mm}

For fast ablation experiments, we only train coarse models on both DTU and RealEstate10K datasets. Since the full RealEstate10K dataset is very large, which would take significant compute to train all the ablations, we use the subset provided by AttnRend~\cite{du2023learning} for training and evaluation.

\boldstartspace{Volume Orientation}. One key difference of our method with previous approaches is that we construct the volume in the \emph{target view frustum}, unlike the popular reference view-based volume in multi-view stereo related methods~\cite{chen2021mvsnerf,johari2022geonerf}. Such a volume representation essentially aggregates information from multi-view input images, where the reference view volume would suffer from information loss since the regions outside its epipolar lines are less likely to be sampled. This issue would be more pronounced for large baselines, as illustrated in Fig.~\ref{fig:visual_ablation_target_volume}. In Table~\ref{tab:ablation}, we can observe a 1dB PSNR performance drop by replacing our target view volume with the reference view volume on the large baseline RealEstate10K dataset. Even on the small baseline DTU dataset, our target view volume is still better since it maximizes the information usage from input views.

\begin{figure}[!t]
\centering
\vspace{1.5mm}
\includegraphics[width=\linewidth]{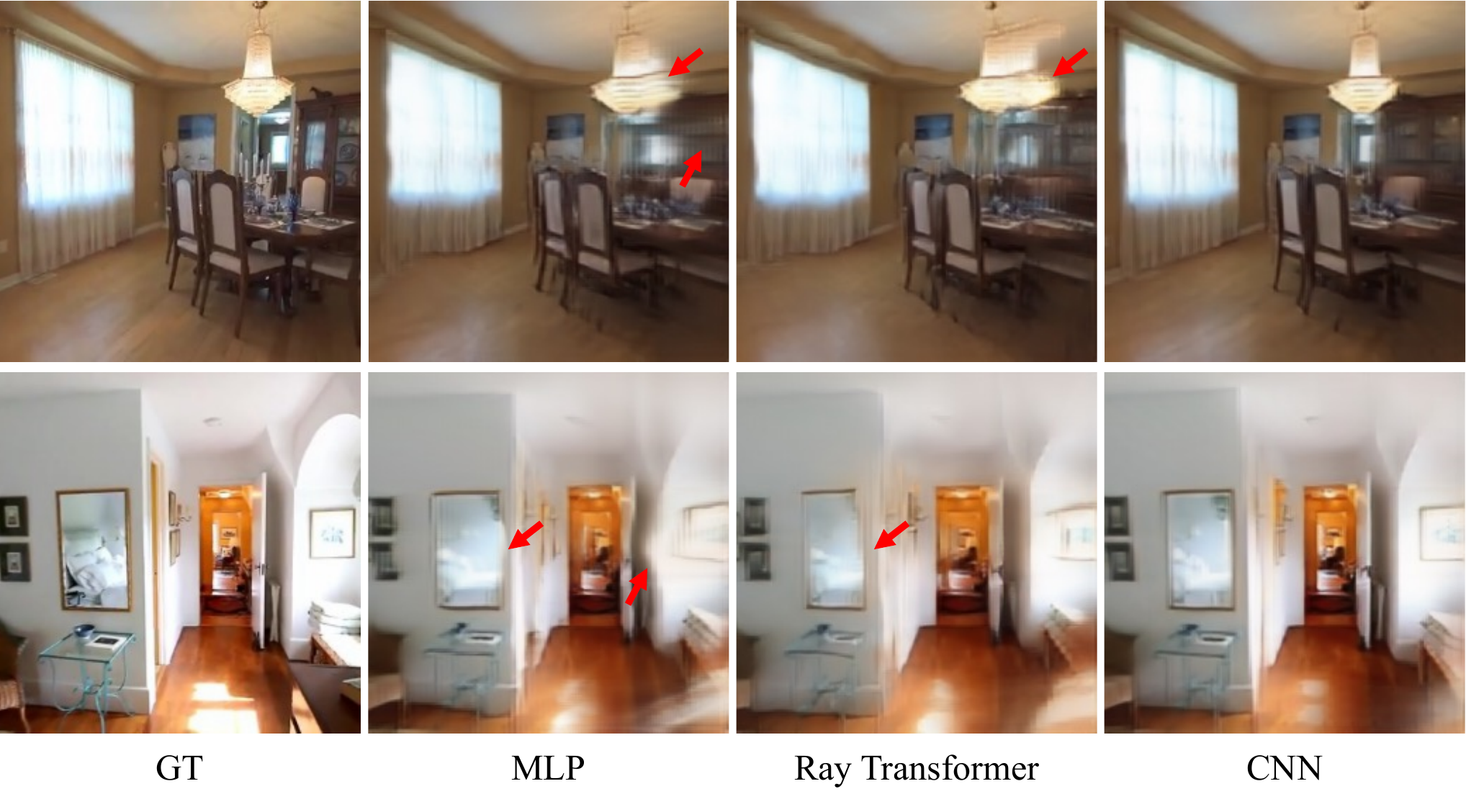}
\vspace{-8mm}
\caption{\textbf{CNN \vs MLP \vs Ray Transformer}. Our CNN-based decoder yields sharper structures thanks to its context modeling abilities in both spatial and depth dimensions.
}
\label{fig:ablation_cnn_vs_mlp}
\vspace{-4mm}
\end{figure}

\vspace{-1.5mm}
\boldstartspace{Volume Elements}. We construct the volume using information from the sampled colors, features, and features' cosine similarities, they are concatenated together as the volume elements. We ablate different inputs in Table~\ref{tab:ablation} to understand their roles. It's interesting to see that for the large baseline setting, the features are more important than the cosine similarities, while cosine similarities become more important for small baselines. This could be expected since the feature matching information would be less reliable for large baselines due to insufficient scene overlap. It's also worth noting that for both small and large baselines, the sampled colors also contribute to the final performance. The image color provides valuable cues for predicting the 3D points' color, thus leading to improved performance.

\vspace{-1.5mm}
\boldstartspace{Radiance Decoder}. We compare different approaches to predict the radiance field from the target view frustum volume in Table~\ref{tab:ablation}. It can be observed that the spatial context modeled by the 2D CNN is more important to the final performance than that on the depth dimension. Previous methods like IBRNet~\cite{wang2021ibrnet} and GNT~\cite{Varma2023ICLR} use Ray Transformer to only model the context information on the depth dimension. However, we observe in our experiments that the Ray Transformer not always produces better results than MLP (see Fig.~\ref{fig:ablation_cnn_vs_mlp}), while our (2+1)D CNN is consistently better. We also compare with the straightforward 3D CNN decoder and the performance is similar. However, 3D CNN has more than 50\% more parameters than our (2+1)D CNN, and thus we choose to use the (2+1)D CNN for better efficiency.
The visual comparison in Fig.~\ref{fig:ablation_cnn_vs_mlp} illustrates that our CNN-based decoder produces clearly better structures.

\section{Conclusion}

We present MuRF, a feed-forward method to sparse view synthesis from multiple different baseline settings. Key to our approach are the proposed target view frustum volume and CNN-based radiance field decoder. We achieve state-of-the-art performance on various evaluation settings, demonstrating the general applicability of our method.

\boldstartspace{Limitation and Discussion}. 
Currently, large-scale scene-level datasets are still scarce, we expect more diverse training data would improve our performance further. Besides, our model currently assumes known camera parameters and static scenes, extending our method to pose-free scenarios and dynamic scenes could be interesting future directions.

\boldstartspace{Acknowledgements}. We thank Shaofei Wang, Zehao Yu and Stefano Esposito for the insightful comments on the early draft of this work. We thank Tobias Fischer and Kashyap Chitta for the constructive discussions. We thank Takeru Miyato for the help with the RealEstate10K dataset. Andreas Geiger was supported by the ERC Starting Grant LEGO-3D (850533) and the DFG EXC number 2064/1 - project number 390727645.

{
    \small
    \bibliographystyle{ieeenat_fullname}
    \bibliography{main}
}

\clearpage

\section*{Appendix}
\renewcommand{\thesection}{\Alph{section}}
\renewcommand{\thetable}{\Alph{table}}
\renewcommand{\thefigure}{\Alph{figure}}
\setcounter{section}{0}
\setcounter{table}{0}
\setcounter{figure}{0}

In this document, we provide high-resolution rendering results, more visual results and more implementation details. We invite the readers to our project page \url{https://haofeixu.github.io/murf/} for more video results.

\section{High-Resolution Rendering}

Our MuRF is developed with the target view frustum volume representation. The volume resolution of the coarse model is $\frac{H}{8} \times \frac{W}{8} \times D_1 \times C_1$, and the fine model is $H \times W \times D_2 \times C_2$ for $H \times W$ image resolution to render, where $D_1=64$ and $D_2=16$ are the numbers of sampling points on each ray, and $C_1=128$ and $C_2=16$ are the volume's feature dimensions. Such resolutions are usually acceptable for typical image resolutions (\eg, $512 \times 512$) on general hardware. Should the memory consumption become a bottleneck for high-resolution images, we can always switch to the patch-based rendering strategy. More specifically, we first split the volume's first two spatial dimensions to a total number of $P \times P$ overlapping patches, and then render each patch independently. Finally, we merge all the patch results to a full image, where the overlapping regions are combined with simple averaging. Such a patch-based rendering strategy enables our method to scale to virtually arbitrary image resolutions.

\begin{figure*}[t]
\thisfloatpagestyle{empty}

    \centering
    \vspace{-20mm}

    \begin{subfigure}{\linewidth}
        \includegraphics[width=\linewidth]{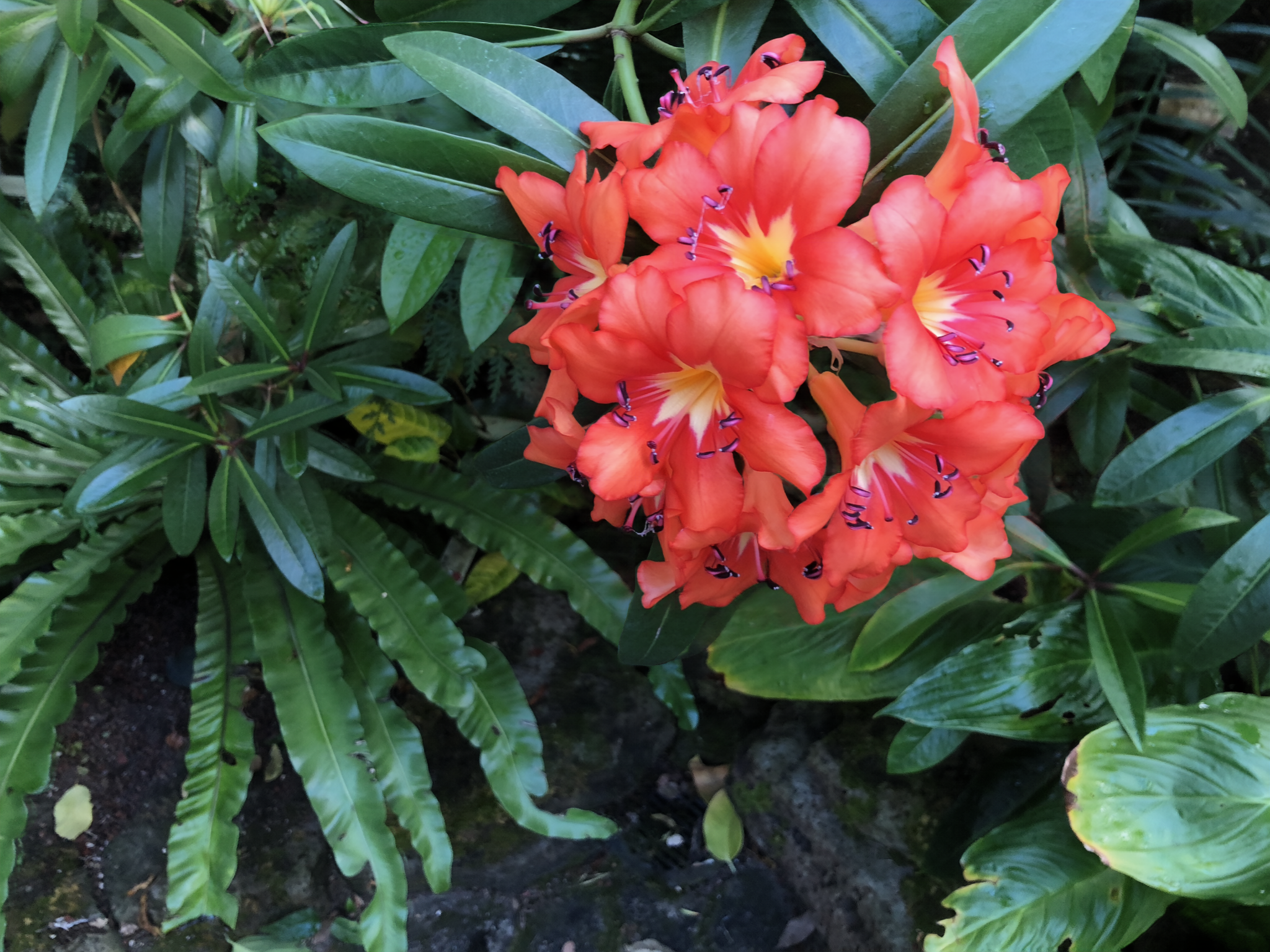}
    \end{subfigure}

    \begin{subfigure}{\linewidth}
        \centering
        \includegraphics[width=\linewidth]{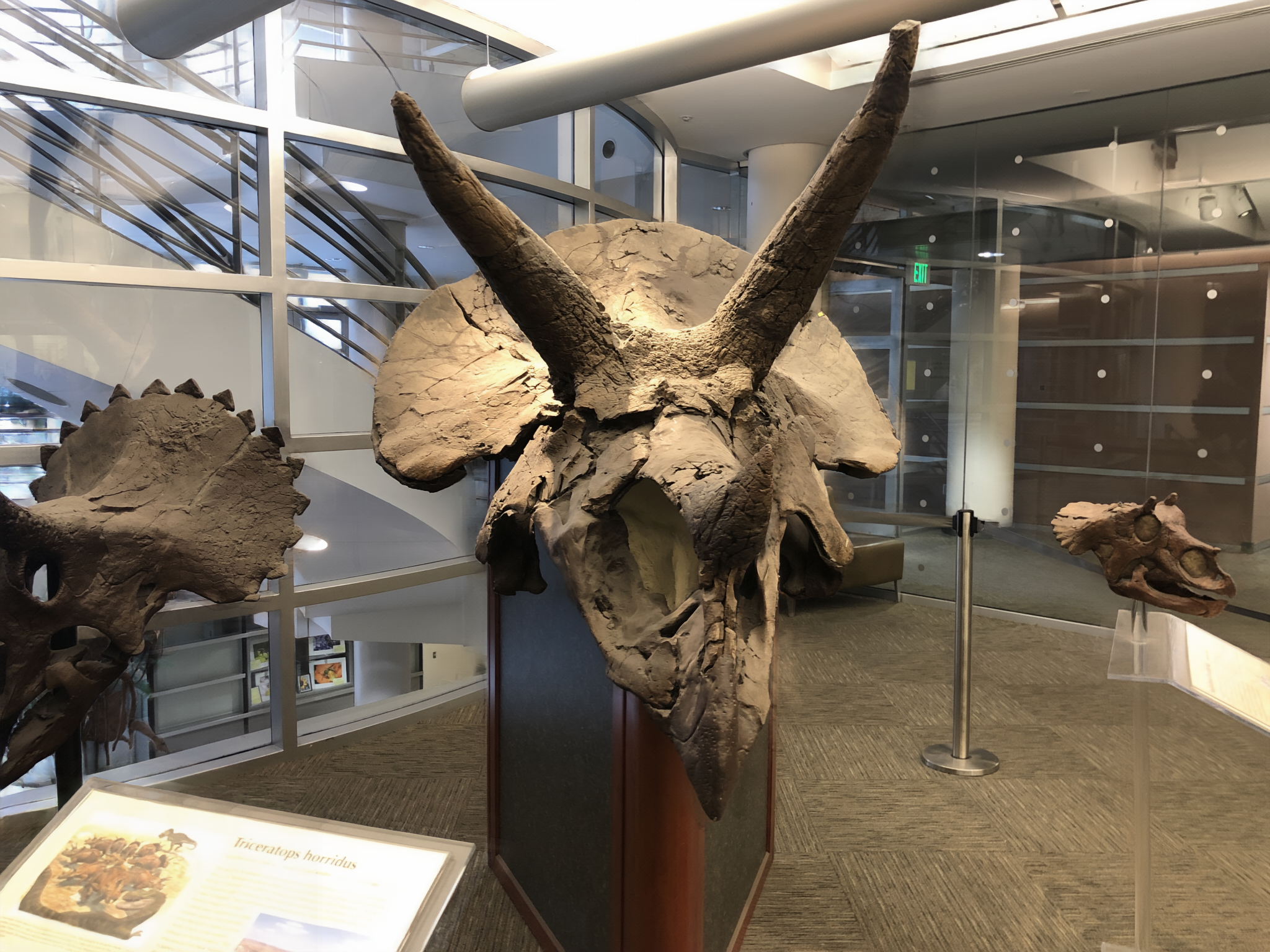}
    \end{subfigure}
    \vspace{-4mm}

    \caption{\textbf{$\bm{1536 \times 2048}$ resolution renderings on the LLFF dataset}.
    }
    \label{fig:highres}
    \vspace{-5mm}
\end{figure*}

In Fig.~\ref{fig:highres}, we show $1536 \times 2048$ resolution rendering results on the LLFF~\cite{mildenhall2020nerf} dataset, where the results are obtained by splitting the full resolution volume to 16 ($4 \times 4$) overlapping patches.

\section{More Visual Results}

\begin{figure}[t]
\centering
\includegraphics[width=\linewidth]{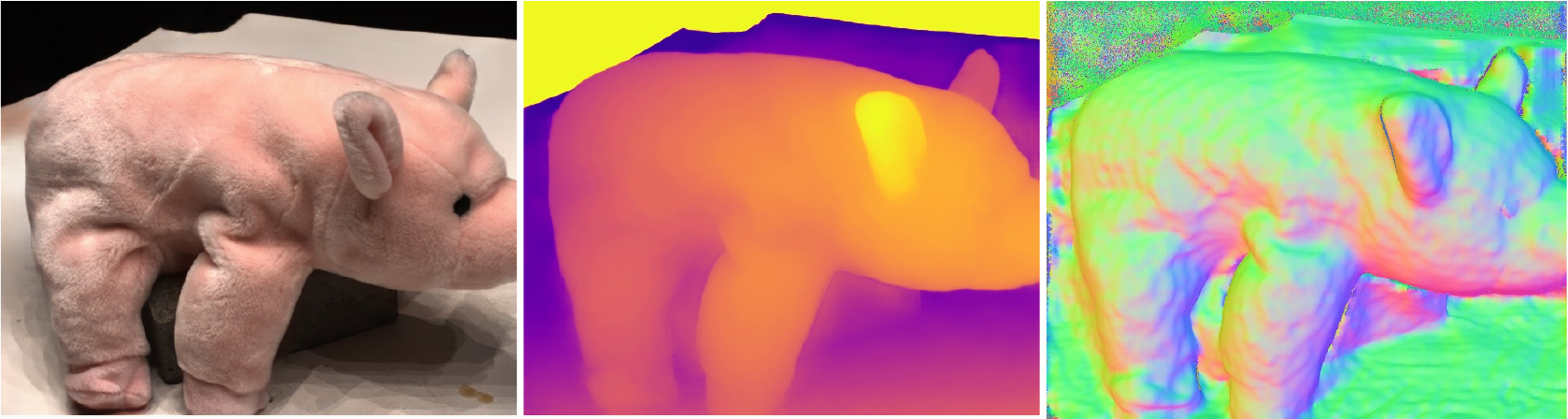}
\vspace{-7mm}
\caption{\textbf{Rendered image, depth and normal from 3 views}.
}
 \label{fig:depth_normal}
\vspace{-6mm}
\end{figure}

\boldstartspace{Geometry Visualization}. In Fig.~\ref{fig:depth_normal}, we show the rendered depth and normal maps from our model, which indicates that our model learned 3D concepts from pure RGB images.

\begin{figure*}[t]
\thisfloatpagestyle{empty}
\centering
\includegraphics[width=\linewidth]{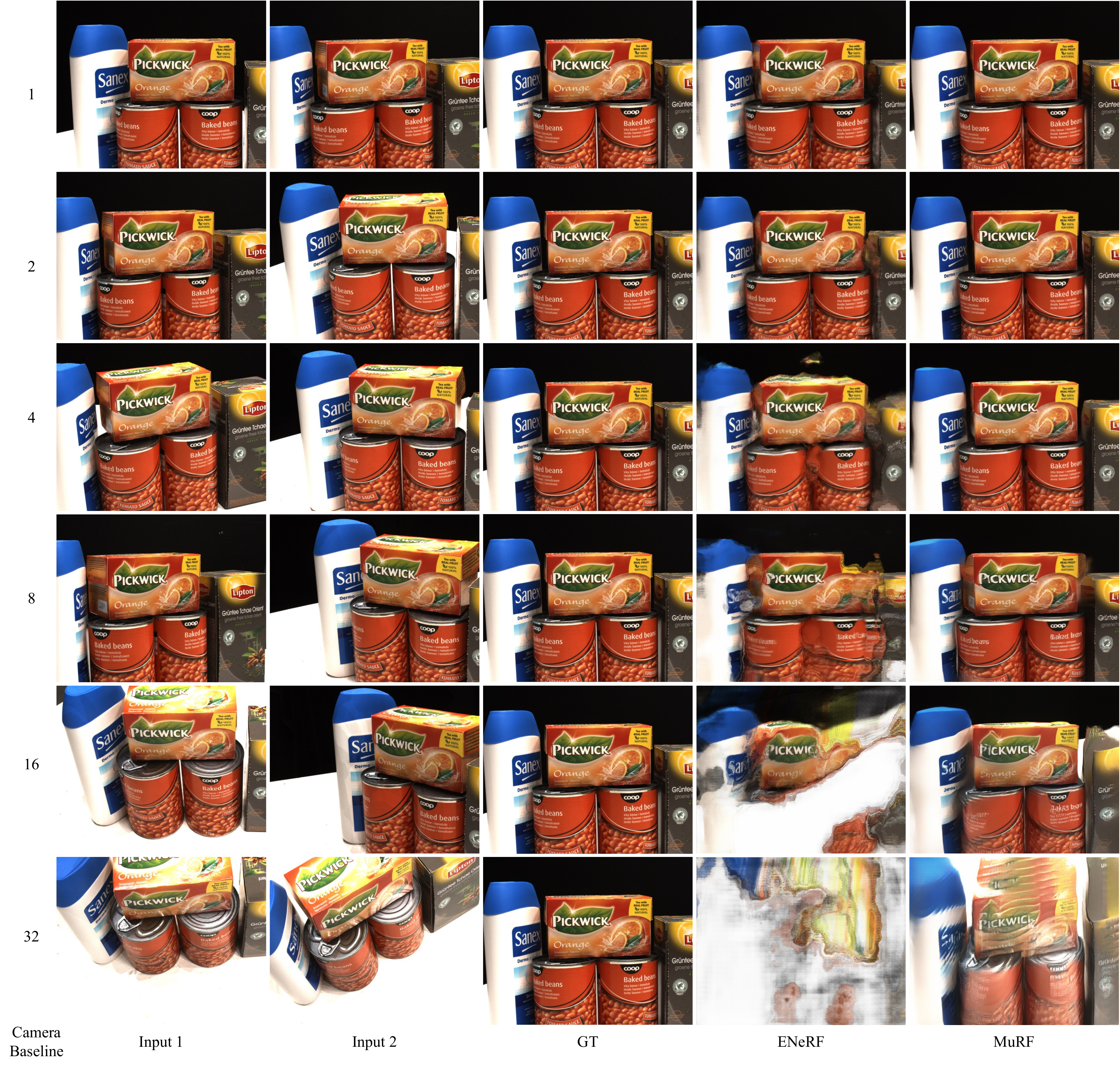}
\caption{\textbf{Results of different camera baselines on DTU}. Our MuRF consistently outpeforms previous state-of-the-art small baseline method ENeRF~\cite{lin2022efficient}, and the performance gap becomes larger for larger baselines.
}
 \label{fig:dtu_different_baselines}
\end{figure*}

\begin{figure*}[t]
\vspace{-4mm}

\centering
\includegraphics[width=\linewidth]{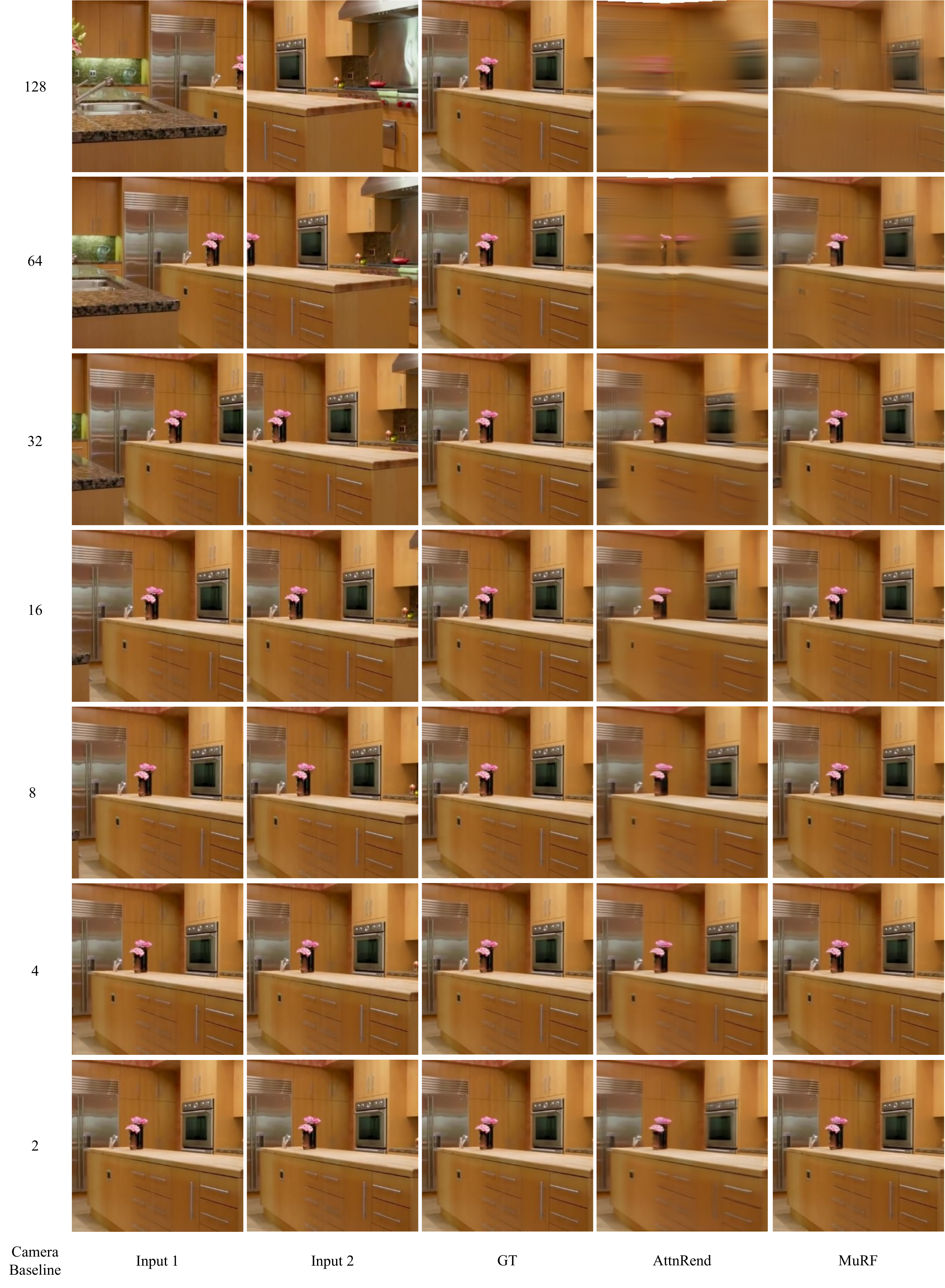}
\vspace{-8mm}

\caption{\textbf{Results of different camera baselines on RealEstate10K}. Our MuRF consistently outpeforms previous state-of-the-art large baseline method AttnRend~\cite{du2023learning}, and our method gains larger improvement for smaller baselines than AttnRend.
}
 \label{fig:realestate_different_baselines}
\end{figure*}

\boldstartspace{Different Camera Baselines}. In Fig.~\ref{fig:dtu_different_baselines}, we show the visual comparison results with previous state-of-the-art small baseline method ENeRF~\cite{lin2022efficient} on the DTU dataset. Our MuRF consistently outperforms ENeRF in different baselines, and the performance gap becomes larger for larger baselines. In Fig.~\ref{fig:realestate_different_baselines}, we show the visual comparison results with previous state-of-the-art larger baseline method AttnRend~\cite{du2023learning} on the RealEstate10K~\cite{DBLP:journals/tog/ZhouTFFS18} dataset. Our MuRF consistently outperforms AttnRend in different baselines. Our method also gains larger improvement for smaller baselines than AttnRend, and our renderings are sharper, while AttnRend's results tend to be blurry.

\begin{figure*}[t]
\thisfloatpagestyle{empty}

    \centering

    \begin{subfigure}{\linewidth}
        \includegraphics[width=\linewidth]{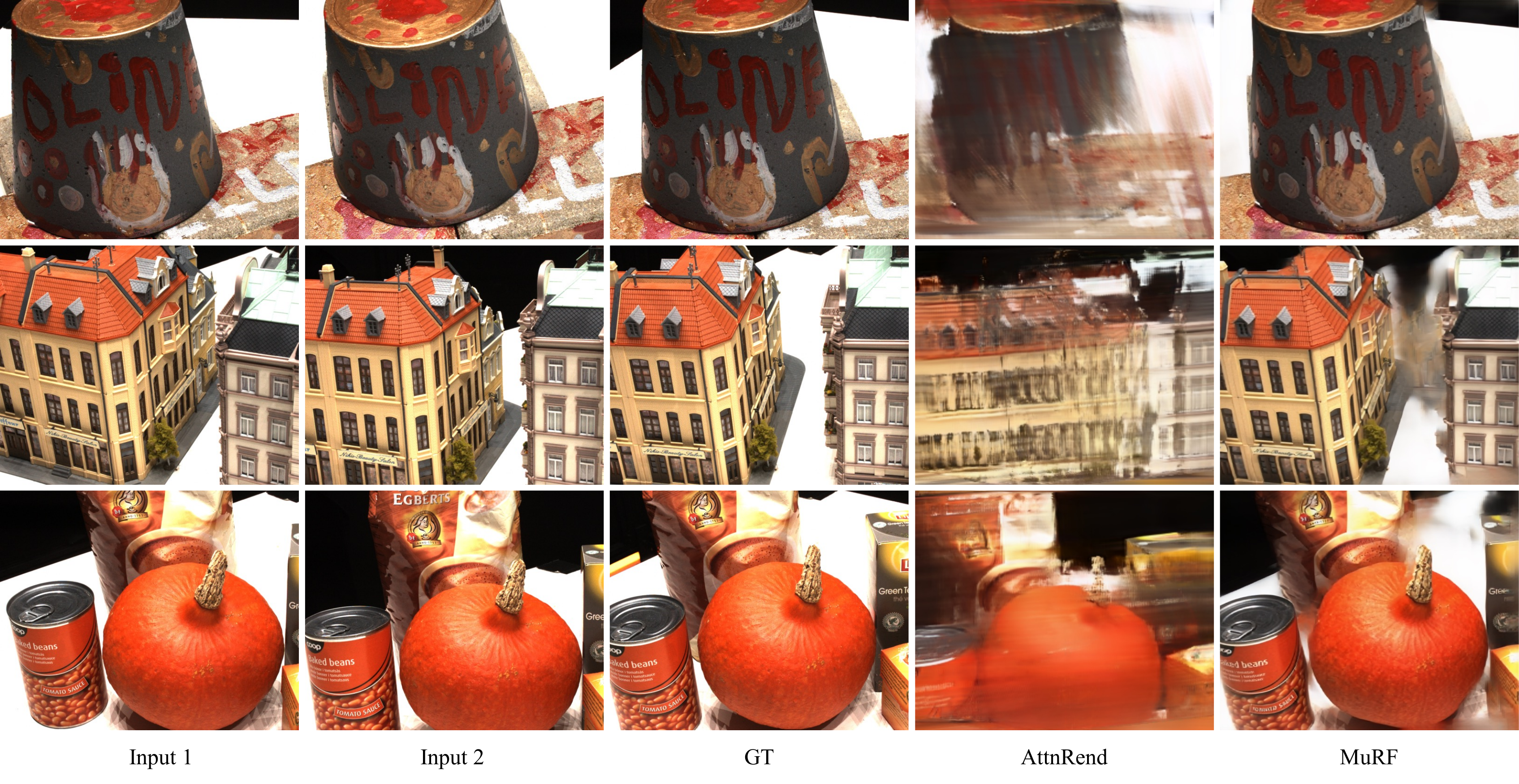}
    \end{subfigure}

    \vspace{-1.7em} %

    \begin{subfigure}{\linewidth}
        \centering
        \includegraphics[width=\linewidth]{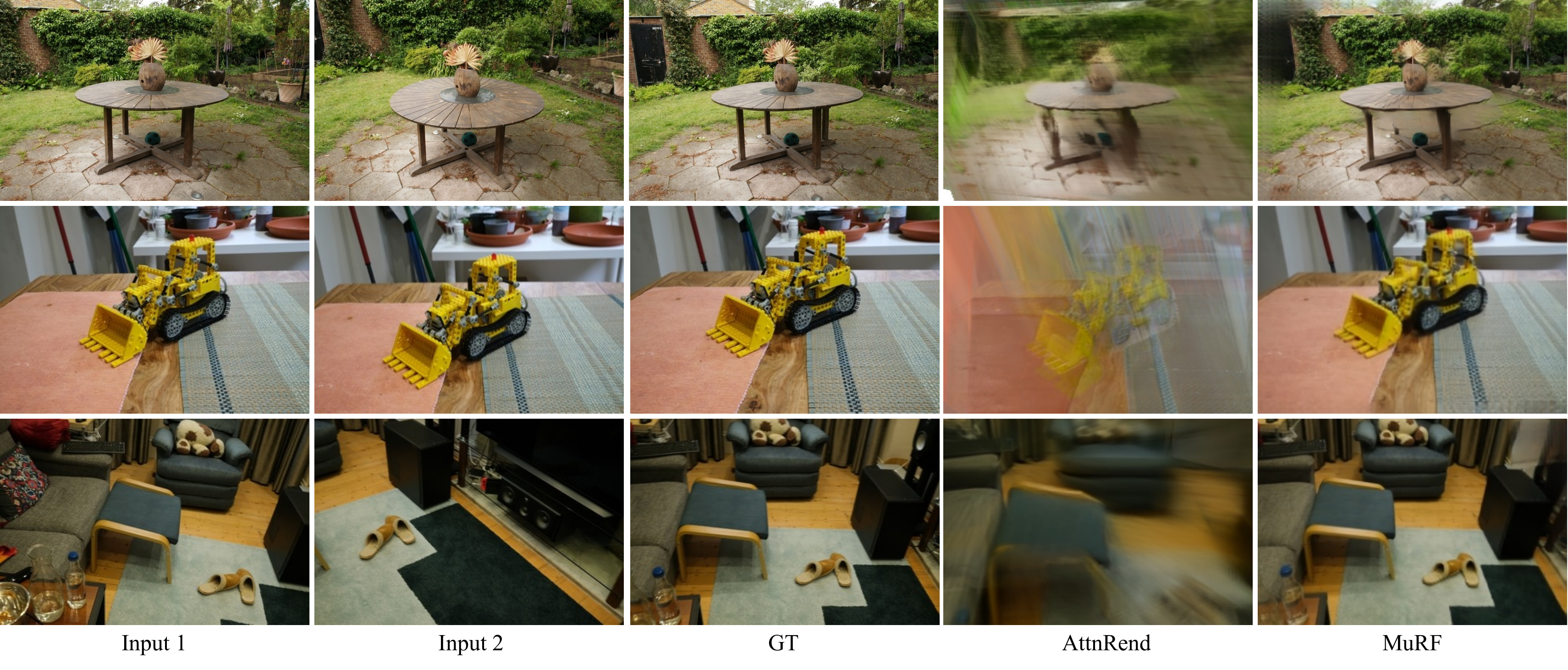}
    \end{subfigure}
    \vspace{-6mm}

    \caption{\textbf{Generalization on DTU and Mip-NeRF 360 dataset with the model trained on RealEstate10K}.
    }
    \label{fig:gen_dtu_mipnerf360}
    \vspace{-5mm}
\end{figure*}

\boldstartspace{Cross-Dataset Generalization}. In Fig.~\ref{fig:gen_dtu_mipnerf360}, we show the cross-dataset generalization results on DTU~\cite{jensen2014large} and Mip-NeRF 360~\cite{barron2022mip} dataset with the model trained on RealEstate10K~\cite{DBLP:journals/tog/ZhouTFFS18}. Our MuRF outperforms AttnRend~\cite{du2023learning} by significant margins.

\section{More Implementation Details}

Following MatchNeRF~\cite{chen2023explicit}, we initialize our multi-view Transformer encoder with GMFlow~\cite{xu2022gmflow} pre-trained weights. The learning rates of the image encoder and the radiance field decoder are $5 \times 10^{-5}$ and $5 \times 10^{-4}$, respectively. The details on each specific experiment are presented below. We will release all the code and models to ease reproduction.

\boldstartspace{DTU}. The image resolution of the DTU dataset is $512 \times 640$. We train our coarse model for 20 epochs on eight RTX A6000 GPUs with a random crop size of $384 \times 512$. The batch size is 8. The performance of the coarse model on the DTU test set is PSNR: 27.19, SSIM: 0.925, LPIPS: 0.120. We then train the fine model with the coarse model frozen. The fine model is trained for 12 epochs with a random crop size of $256 \times 384$, and the batch size is 8.

\boldstartspace{RealEstate10K}. We use the image resolution of $256 \times 256$ on the RealEstate10K dataset following AttnRend~\cite{du2023learning}. For this resolution, we use $4 \times$ subsampling when constructing the volume and no additional hierarchical sampling is used. We train the model for 50 epochs on three A100 GPUs with a random crop size of $224 \times 224$, and the batch size is 6.

\boldstartspace{LLFF}. The testing image resolution of the LLFF~\cite{mildenhall2020nerf} dataset is $756 \times 1008$, and the training data consists of several mixed datasets following previous works~\cite{wang2021ibrnet,suhail2022generalizable}. We train models with different numbers (2, 6, and 10) of input views to compare with previous methods. The 2-view model is trained with a random crop size of $384 \times 512$, and the 6-view and 10-view models are trained with $256 \times 384$ random crops. The 2-view and 6-view models are trained on eight RTX A6000 GPUs, and the 10-view model is trained on two A100 GPUs.

\boldstartspace{Ablations}. For ablation experiments on the DTU and RealEstate10K datasets in the main paper, we only train the coarse models without the hierarchical sampling. All ablations are trained two RTX 3090 GPUs. The DTU models are trained for 20 epochs with a random crop size of $256 \times 416$, and the RealEstate10K models are trained for 80 epochs with the full $256 \times 256$ resolution.

\end{document}